\documentclass[sigconf]{acmart}

\usepackage{graphicx}
\usepackage{amsmath}
\usepackage{multirow}
\usepackage{balance}
\usepackage{algorithm}
\usepackage{algorithmic}

\AtBeginDocument{%
  \providecommand\BibTeX{{%
    \normalfont B\kern-0.5em{\scshape i\kern-0.25em b}\kern-0.8em\TeX}}}

\copyrightyear{2021}
\acmYear{2021}
\setcopyright{acmcopyright}
\acmConference[MM '21]{Proceedings of the 29th ACM International Conference on Multimedia}{October 20--24, 2021}{Virtual Event, China}
\acmBooktitle{Proceedings of the 29th ACM International Conference on Multimedia (MM '21), October 20--24, 2021, Virtual Event, China}
\acmPrice{15.00}
\acmDOI{10.1145/3474085.3475232}
\acmISBN{978-1-4503-8651-7/21/10}

\acmSubmissionID{mfp0402}

\begin{document}
\fancyhead{}

\title{Few-shot Unsupervised Domain Adaptation with Image-to-class Sparse Similarity Encoding}

\author{Shengqi Huang}
\affiliation{
  \institution{Nanjing Normal University}
  \city{Nanjing}
  \country{China}
}
\email{huangshengqi@njnu.edu.cn}

\author{Wanqi Yang}\authornote{Corresponding author}
\affiliation{
  \institution{Nanjing Normal University}
  \city{Nanjing}
  \country{China}}
\email{yangwq@njnu.edu.cn}

\author{Lei Wang}
\affiliation{
  \institution{University of Wollongong}
  \city{Wollongong}
  \country{Australia}
}
\email{leiw@uow.edu.au}

\author{Luping Zhou}
\affiliation{
 \institution{University of Sydney}
 \city{Sydney}
 \country{Australia}}
\email{luping.zhou@sydney.edu.au}

\author{Ming Yang}
\affiliation{
  \institution{Nanjing Normal University}
  \city{Nanjing}
  \country{China}}
\email{myang@njnu.edu.cn}

\begin{abstract}
  This paper investigates a valuable setting called few-shot unsupervised domain adaptation (FS-UDA), which has not been sufficiently studied in the literature. In this setting, the source domain data are labelled, but with few-shot per category, while the target domain data are unlabelled. To address the FS-UDA setting, we develop a general UDA model to solve the following two key issues: the few-shot labeled data per category and the domain adaptation between support and query sets. Our model is general in that once trained it will be able to be applied to various FS-UDA tasks from the same source and target domains. Inspired by the recent local descriptor based few-shot learning (FSL), our general UDA model is fully built upon local descriptors (LDs) for image classification and domain adaptation. By proposing a novel concept called similarity patterns (SPs), our model not only effectively considers the spatial relationship of LDs that was ignored in previous FSL methods, but also makes the learned image similarity better serve the required domain alignment. Specifically, we propose a novel \textbf{IM}age-to-class sparse \textbf{S}imilarity \textbf{E}ncoding (\textbf{IMSE}) method. It learns SPs to extract the local discriminative information for classification and meanwhile aligns the covariance matrix of the SPs for domain adaptation. Also, domain adversarial training and multi-scale local feature matching are performed upon LDs. Extensive experiments conducted on a multi-domain benchmark dataset \emph{DomainNet} demonstrates the state-of-the-art performance of our IMSE for the novel setting of FS-UDA. In addition, for FSL, our IMSE can also show better performance than most of recent FSL methods on \emph{miniImageNet}.
\end{abstract}



\maketitle

\section{Introduction}\label{intro}
Unsupervised domain adaptation (UDA) leverages sufficient labeled data in a source domain to classify unlabeled data in a target domain \cite{ADDA,MCD,UDA1}. Since the source domain usually has a different distribution from the target domain, most existing UDA methods \cite{hierarchical_UDA,MCD,UDAlabel18} have been developed to reduce the domain gap between both domains. However, in real cases, when the source domain suffers from high annotation cost or the limited access to labeled data, the labeled data in source domain could become seriously insufficient. This makes the learning of a UDA model more challenging. On the other hand, existing UDA methods usually train a specific UDA model for a certain task with given categories, and thus cannot be efficiently updated to classify unseen categories in a new task. Recently, the need of handling few-shot labeled source domain data and having a general UDA model that can quickly adapt to many new tasks has been seen in a number of real applications.

For example, in the \emph{online shopping}, sellers generally show a \emph{few photos} professionally taken by photographers for each product (\emph{as source domain}), while buyers usually search for their expected products by using the photos directly taken by their cellphones (\emph{as target domain}). Since different buyers usually submit the images of products with different categories and types, the online shopping system needs to have a general domain adaptation model to classify the types of these products, whatever their categories are. Also, we take another example. The database of \emph{ID card} usually only stores \emph{one photo} for each person and these ID photos are captured in photo studios with well controlled \emph{in-door} lighting conditions (\emph{as source domain}). However, in many cases, an intelligence system needs to identity these persons by matching with their facial pictures captured by smartphones, ATM cameras, or even surveillance cameras in open environments (\emph{as target domain}). 

This situation motivates us to focus on a valuable but insufficiently studied setting, few-shot unsupervised domain adaptation (FS-UDA). A FS-UDA task includes a support set\footnote{Following the terminology of few-shot learning (FSL), we call the training and testing sets in a classification task as the support and query sets, respectively, in the paper.} from the source domain and a query set from the target domain. The support set contains few labeled source domain samples per category, while the query set contains unlabeled target domain samples to be classified. The goal is to train a general model to handle various FS-UDA tasks from the same source and target domains. These tasks realize classification of different category sets in the target domain.

Our solution to this FS-UDA setting is inspired by few-shot learning (FSL). Recent metric based FSL methods \cite{matchingnet,protonet,DN4,DeepEMD} usually learn a feature embedding model for \emph{image-to-class} measurement, and meanwhile perform episodic training\footnote{\emph{Episodic training} is to train a model on large amounts of episodes. For each episode, a few samples are randomly sampled in the auxiliary dataset to train the model, instead of the whole dataset.} to make the model generalized to new tasks without any retraining or fine-tuning process. Considering the domain gap between the support and query sets in the proposed setting of FS-UDA, we tactically perform episodic training on an auxiliary dataset that contains the data from both domains, aiming to learn a general and domain-shared feature embedding model. For each episode, a few samples from both domains are randomly sampled from the auxiliary dataset to train the model. 

Also, for efficient \emph{image-to-class} measurement, several recent methods \cite{DN4,DeepEMD,ADM} leveraged local descriptors\footnote{A local descriptor (LD) is referred to a $C$-dimensional vector along with the channels at a certain pixel in the embedded feature map, where $C$ is the channel number.} (LDs) of images, because they can better provide discriminative and transferable information across categories, which can be important cues for image classification in FSL \cite{DeepEMD}. These methods first measured the similarities (or distances) between the LDs of a \emph{query image} and a \emph{support class}\footnote{A \emph{support class} is referred to a class in the support set, and a \emph{query image} represents an image in the query set.} by using \emph{cosine} \cite{DN4}, KL divergence \cite{ADM} or EMD distance \cite{DeepEMD}, and then combined these similarities to produce a single \emph{image-to-class} similarity score for classification. We believe that for two LDs that are spatially adjacent and therefore highly likely similar to each other, their similarity values to other LDs should be close and change in a smooth manner. However, the above FSL methods did not effectively take the spatial relationship of the LDs into account, and could result in noisy or non-smooth similarities \emph{w.r.t.} the spatial dimension (\emph{e.g.} caused by outliers, occlusion or background clutter in images). Moreover, considering the key role of \emph{image-to-class} similarities for classification, we leverage them to align the source and target domains. Nevertheless, merely having a single similarity score as produced in the above methods cannot effectively serve the domain alignment required in our FS-UDA.

\begin{figure}[t]
\vspace{-0.8cm}
\begin{center}
\includegraphics[width=0.96\linewidth]{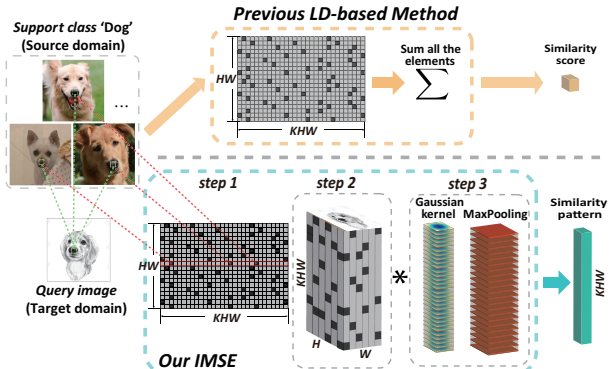}
\end{center}
\vspace{-0.9cm}
\caption{\small{Illustration of encoding a similarity pattern in our IMSE, compared with a single similarity score produced in previous LD-based FSL methods (\emph{e.g.} DN4 \cite{DN4}, ADM \cite{ADM}, DeepEMD \cite{DeepEMD}). For a similarity pattern, we first calculate the similarity matrix between LDs of a \emph{query image} to a \emph{support class}, and only maintain the top-$k$ similarities for each row. Then, the similarity matrix is reshaped into a 3D map whose spatial sizes are the same as those of the \emph{query image}. Afterwards, we conduct a Gaussian filter and max-pooling operation to smooth and aggregate the similarity of the spatially-adjacent LDs of the \emph{query image}, and thus generate an \emph{image-to-class} SP finally. Here, $H$ and $W$ denote the height and weight of the feature map, and $K$ denotes the number of images per \emph{support class}.}}
\label{fig1}
\vspace{-0.4cm}
\end{figure}

Therefore, instead of computing a single similarity score, we design a novel \emph{image-to-class} similarity pattern (SP) vector that contains the similarities of a \emph{query image} to all LDs of a \emph{support class}. We design a Gaussian filter and a max-pooling module on the spatial dimension of \emph{query images}, which can smooth the similarities of spatially-adjacent LDs and extract the prominent similarity. Thus, SP provides a fine-grained and patterned measurement for classification. Also, the SPs of \emph{query images} can be leveraged as their features for domain alignment. To this end, we align the covariance matrix of the SPs between domains, because covariance matrix can naturally capture the underlying data distribution \cite{covnet}. For further domain alignment, we also perform domain adversarial training and multi-scale matching on LDs. In doing so, we propose a novel method, \textbf{IM}age-to-class sparse \textbf{S}imilarity \textbf{E}ncoding (IMSE), which learns \emph{image-to-class} similarity patterns to realize effective classification and domain adaptation. Figure \ref{fig1} illustrates the encoding process of SPs. Our contributions can be summarized as:

\begin{enumerate}
    \item \textbf{A novel method for the new setting of FS-UDA.} We propose a novel method IMSE for a valuable but insufficiently studied setting FS-UDA, which learns a domain-shared feature embedding model and adapts this general model to handle various new tasks. 
    \item \textbf{A novel \emph{image-to-class} measurement between source and target domains.} We propose \emph{image-to-class} similarity patterns that consider the spatial information of local descriptors. Moreover, we leverage the similarity patterns and local descriptors to further align the domains.
    \item \textbf{Effective performance.} Extensive experiments are conducted on a mutli-domain benchmark dataset \emph{DomainNet}, showing the state-of-the-art performance of our method for FS-UDA. In addition, for FSL, our IMSE also shows better performance than most recent FSL methods on \emph{miniImageNet}.
\end{enumerate}

\begin{table*}
\centering
\small
\caption{The main differences between the settings of UDA, FSL, cross-domain FSL and our FS-UDA.}\label{diff}
\vspace{-0.4cm}
\renewcommand{\arraystretch}{0.9}
\setlength{\tabcolsep}{8pt}
\begin{tabular}{c|c|c|c|c}\hline\hline
\multirow{2}{*}{\textbf{Settings}} & \textbf{Domain gap between} & \textbf{No available labels} & \textbf{Few-shot labeled samples} & \textbf{Handling many tasks with}\\
& \textbf{the support and query sets} & \textbf{in the target domain} & \textbf{per category}& \textbf{different sets of categories} \\\hline
\textbf{UDA} & \checkmark & \checkmark &  & \\
\textbf{FSL} &  & & \checkmark & \checkmark \\
\textbf{Cross-domain FSL} &  &  & \checkmark & \checkmark (tasks from new domains)\\
\textbf{FS-UDA} (ours)  & \checkmark & \checkmark & \checkmark & \checkmark \\\hline\hline
\end{tabular}
\vspace{-0.3cm}
\end{table*}

\section{Related Work}
\textbf{Unsupervised domain adaptation.} The UDA setting aims to reduce domain gap and leverage sufficient labeled source domain data to realize classification in target domain. Many UDA methods \cite{UDA1, Long2015} are based on maximum mean discrepancy to minimize the feature difference across domains. Besides, adversarial training is widely used \cite{ADDA, DANN} to tackle domain shift. ADDA \cite{ADDA} performs domain adaptation by using an adversarial loss and learning two domain-specific encoders. The maximum classifier discrepancy method (MCD) \cite{MCD} applies task-specific classifiers to perform adversarial learning with the feature generator. Besides, several methods \cite{hierarchical_UDA, UDAlabel20, UDAlabel18} train the classifier in target domain by leveraging their pseudo-labels. SDRC \cite{tang1} learns clustering assignment of target domain and leverage them as the pseudo-labels to compute classification loss. In sum, these UDA methods require sufficient labeled source domain data for domain alignment and classification, but would not work when encountering the issues of scarce labeled source domain and task-level generalization that exist in FS-UDA.

\textbf{Few-shot learning.} 
Few-shot learning has become one of the most popular research topics recently, which aims to quickly adapt the model to new few-shot tasks. It contains two main branches: \emph{optimization-based} and \emph{metric-based} approaches. \emph{Optimization-based} methods \cite{maml,opt1, R2D2}  leverage a meta learner over the auxiliary dataset to learn a general initialization model. When handling new tasks, the initialization model can be quickly fine-tuned to adapt to new tasks. \emph{Metric-based} methods \cite{protonet,matchingnet, DN4, FEAT} aim to learn a generalizable feature embedding metric space by episodic training on the auxiliary dataset \cite{matchingnet}. This metric space can be directly applied to new few-shot tasks without any fine-tuning or re-training. Our method is related to \emph{metric-based} methods. Classically, ProtoNet \cite{protonet} learns the class prototypes in the support set and classifies the \emph{query images} by calculating the maximum similarity to these prototypes. CAN \cite{CAN} learns the cross-attention map between \emph{query images} and \emph{support classes} to highlight the discriminative regions. Due to the local invariance of local descriptors (LDs), several methods \cite{DN4, DeepEMD, denseLD, ADM} leverage large amounts of LDs to solve the FSL problem. For example, DN4 \cite{DN4} classifies \emph{query images} by finding the neighborhoods of their LDs amongs LDs of the \emph{support classes}. Based on LDs, DeepEMD \cite{DeepEMD} measures earth mover's distance of \emph{query images} to the structured prototype of \emph{support classes} for \emph{image-to-class} optimal matching. For performance improvement, recent FSL methods usually pre-train the embedding network \cite{meta-baseline} on auxiliary dataset. Inspired by these methods, we leverage LDs to calculate \emph{image-to-class} similarity patterns (SPs) for classification and domain alignment in our setting. Furthermore, a few recent works focus on the issue of \textbf{cross-domain FSL} in which domain shift exists between data of meta tasks and new tasks. Chen \emph{et al.} \cite{jiabin2} leverage a baseline model to do cross-domain FSL. LFT \cite{jiabin1} designs a feature transformation strategy to tackle the domain shift.

\textbf{Remarks.} Note that \textbf{the setting of FS-UDA is crucially different from the existing settings: UDA \cite{ADDA,MCD}, FSL \cite{DN4,DeepEMD} and cross-domain FSL \cite{jiabin1,jiabin2}}. As shown in Table \ref{diff}, compared with UDA, FS-UDA is to deal with many UDA tasks by leveraging few labeled source domain samples for each. As known, FSL is to train a general classifier by leveraging few-shot training samples per category and adapt the classifier to new tasks. The new tasks for FSL are sampled from the same domain as the tasks for training, while the new tasks for cross-domain FSL could be from new domains. Compared with FS-UDA, FSL and cross-domain FSL are not capable of handling the issues of no available labels in the target domain, and the domain gap between the support and query sets in every task, while these issues will be solved in our FS-UDA. In addition, \textbf{the FS-UDA setting also has key differences from existing few-shot domain adaptation methods \cite{motiian2017few,xu2019d,teshima2020few}}, because i) the ``few-shot'' in the latter is referred to labeled target domain data and ii) they still need sufficient labeled source domain data to work. For the one-shot UDA recently proposed in \cite{luo2020adversarial}, it deals with the case that only one unlabeled target sample is available, but does not require the source domain to be few-shot, which is different from ours. In short, the above methods focus on data or label scarcity in target domain rather than in source domain. Also, they do not consider developing a single model that can generally be adapted to many different UDA tasks. These differences indicate the necessity of the novel FS-UDA setting, for which we will explore a new learning algorithm to address the data scarcity in source domain and the UDA-task-level generalization.

\begin{figure*}
\hsize=\textwidth
\begin{center}
\includegraphics[width=0.88\textwidth]{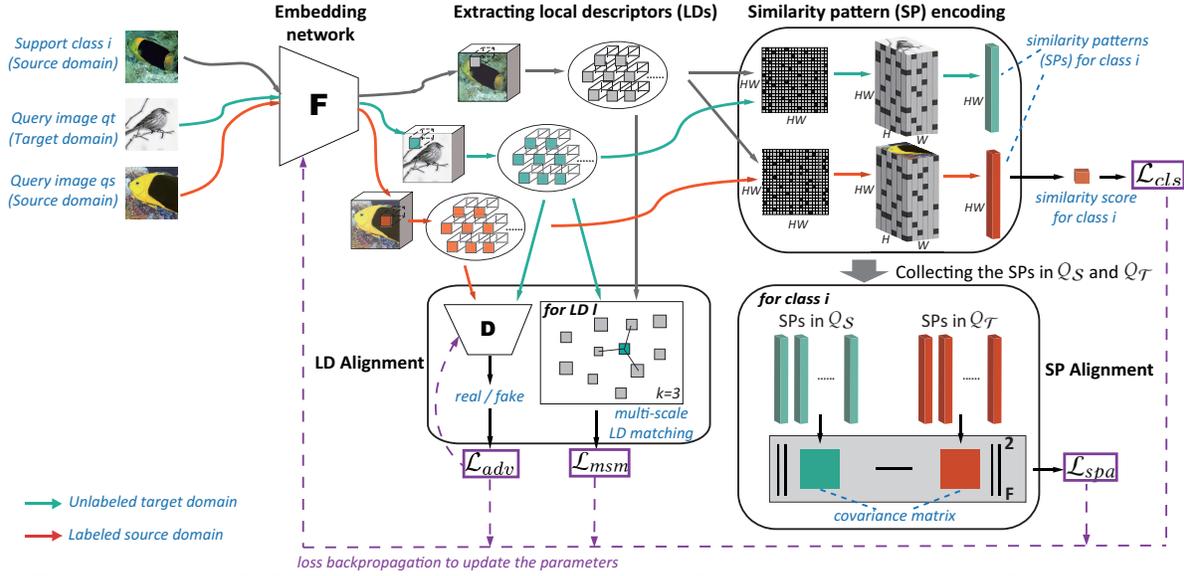}
\end{center}
\vspace{-0.5cm}
   \caption{\small{Illustration of our method performing episodic training for \emph{5-way, 1-shot} UDA tasks. In each episode, the support set $X_{\mathcal{S}}$ contains a photo image per \emph{support class}, and the query sets ($Q_{\mathcal{S}}$ and $Q_{\mathcal{T}}$) contains the photo and sketch \emph{query images}, respectively. \emph{Support class} $i$ and \emph{query images} $qs, qt$ from both domains are first through the feature embedding network $F$ to extract their local descriptors (LDs), followed by similarity pattern encoding module to learn similarity patterns (SPs). Then, we leverage the SPs in $Q_{\mathcal{S}}$ to calculate the similarity of \emph{query images} to the class $i$ for classification loss $\mathcal{L}_{cls}$. Meanwhile, we calculate the covariance matrix of the SPs between $Q_{\mathcal{S}}$ and $Q_{\mathcal{T}}$ for measuring their domain alignment loss $\mathcal{L}_{spa}$. In addition, the LDs from both domains are further aligned by calculating both adversarial training loss $\mathcal{L}_{adv}$ and multi-scale matching loss $\mathcal{L}_{msm}$. Finally, the above losses are backpropagated to update the embedding model $F$.}}
\label{fig3}
\vspace{-0.3cm}
\end{figure*}

\section{Methodology}
\subsection{Problem Definition}\label{3.1}
\textbf{A \emph{N-way, K-shot} UDA task.} Our FS-UDA setting involves two domains in total: a source domain $\mathcal{S}$ and a target domain $\mathcal{T}$. A \emph{N-way, K-shot} UDA task includes a support set $X_{\mathcal{S}}$ from $\mathcal{S}$ and a query set $Q_{\mathcal{T}}$ 
from $\mathcal{T}$. The support set $X_{\mathcal{S}}$ contains $N$ classes and $K$ source domain samples per class. The query set $Q_{\mathcal{T}}$ contains the same $N$ classes as in $X_{\mathcal{S}}$ and $N_q$ target domain samples per class. To classify \emph{query images} in $Q_{\mathcal{T}}$ to the correct class in $X_S$, as a conventional solution, leveraging the insufficient support set $X_{\mathcal{S}}$ to train a specific model from scratch could make the classification inefficient. Therefore, we aim to train a general model that can handle many new \emph{N-way, K-shot} UDA tasks for testing. Table \ref{notation} shows the main symbols used in this paper.

\begin{table}
\centering
\setlength{\tabcolsep}{3pt}
\renewcommand{\arraystretch}{0.88}
\small
\caption{Notations.}\label{notation}
\vspace{-0.4cm}
\begin{tabular}{c|l}
\hline
\textbf{Notations} & \hspace{2cm}\textbf{Descriptions} \\
\hline
$N,K\in \mathbb{R}$ & The number of the classes and samples per class\\
\hline
$\lambda_{s},\lambda_{a},\lambda_{m}\in \mathbb{R}$& Weight parameters of three loss terms in Eq. (\ref{obj_func})\\
\hline
\multirow{2}{*}{$X_{\mathcal{S}},\mathcal{Q}_{\mathcal{S}},\mathcal{Q}_{\mathcal{T}}$}&  Support set of source domain, and query sets of\\
& source domain and target domain \\
\hline
$H,W,C\in \mathbb{R}$ & The height, the width and the number of channels,\\
$l\in \mathbb{R}^{C}$& and local descriptor (LD) vector in the feature map\\
\hline
$L_{q}^{que}\in \mathbb{R}^{HW\times C}$ & LD matrix that consists of LDs of a \emph{query image} $q$,\\
$L_{c}^{sup}\in \mathbb{R}^{KHW\times C}$ & LD matrix that consists of LDs of a \emph{support class} $c$\\
\hline
$M_q^c\in\mathbb{R}^{HW\times KHW}$& 2D and 3D matrices of \emph{cosine} similarities between\\
$M_q^{c'}\in\mathbb{R}^{H\times W\times KHW}$& LDs of a \emph{query image} $q$ and a \emph{support class} $c$\\
\hline
$p_{q}^{c}\in \mathbb{R}^{KHW}$ & Similarity pattern (SP) vectors of a \emph{query image} $q$\\
$p_{q}^{i}\in \mathbb{R}^{HW}$& to a \emph{support class} $c$ and a \emph{support image} $i$\\
\hline
\multirow{2}{*}{$\Sigma_{d}^{i}\in \mathbb{R}^{HW\times HW}$} & Covariance matrix of similarity  patterns from all\\
& \emph{query images} in domain $d$ to a \emph{support image} $i$\\
\hline
$F,D$ & Feature embedding network, domain discriminator\\
\hline
\end{tabular}
\vspace{-0.3cm}
\end{table}

\textbf{Auxiliary dataset and episodic training.} Inspired by FSL, we resort to an auxiliary dataset $D^{\text{aux}}$ and perform episodic training to learn the general model. The auxiliary dataset  $D^{\text{aux}}$ includes labeled source domain data and unlabeled target domain data. Note that, the classes in $D^{\text{aux}}$ are completely different from the testing \emph{N-way, K-shot} UDA tasks. That is, the classes of the testing UDA tasks are unseen during episodic training. We construct large amounts of episodes to simulate the testing UDA tasks for task-level generalization \cite{matchingnet}. Each episode contains a support set $X_{\mathcal{S}}$, a source domain query set $Q_{\mathcal{S}}$ and a target domain query set $Q_{\mathcal{T}}$ sampled from $D^{\text{aux}}$. Specifically, in each episode, we randomly sample $N$ classes and $K+N_q$ source domain samples per class from $D^{\text{aux}}$. The $K+N_q$ samples per class are partitioned into $K$ in $X_{\mathcal{S}}$ and the remaining $N_{q}$ in $Q_{\mathcal{S}}$. Meanwhile, we randomly sample $N*N_{q}$ target domain samples (without considering labels) from $D^{\text{aux}}$ and incorporate them into $Q_{\mathcal{T}}$. Note that the query set $Q_{\mathcal{S}}$ per episode is additional, compared with the testing tasks. This is because the labels of $Q_{\mathcal{T}}$ are unavailable, we instead leverage $Q_{\mathcal{S}}$ to calculate classification loss, and meanwhile use $Q_{\mathcal{S}}$ and  $Q_{\mathcal{T}}$ for domain alignment. 

\textbf{The flowchart of our method.} 
Figure \ref{fig3} illustrates our method to perform episodic training for \emph{5-way, 1-shot} UDA tasks. In each episode, a support set ($X_{\mathcal{S}}$) and two query sets ($Q_{\mathcal{S}}$ and $Q_{\mathcal{T}}$) are first through the feature embedding network $F$ to extract their LDs, followed by similarity encoding module to learn similarity patterns (SPs). Then, we leverage the SPs to calculate the \emph{image-to-class} similarity for classification loss $\mathcal{L}_{cls}$. Meanwhile, we calculate the covariance matrix of the SPs between domains to measure their domain alignment loss $\mathcal{L}_{spa}$. In addition, the LDs from domains are further aligned by calculating both adversarial training loss $\mathcal{L}_{adv}$ and multi-scale matching loss $\mathcal{L}_{msm}$. Finally, the above losses are back-propagated to update the embedding model $F$. After episodic training over all episodes, we employ the learned embedding model $F$ to test many \emph{5-way, 1-shot} new UDA tasks. For each new testing task, the support set $X_{\mathcal{S}}$ is used to classify target domain samples in $Q_{\mathcal{T}}$. Then, we calculate the averaged classification accuracy on these tasks for performance evaluation.

\subsection{Similarity Pattern Encoding}\label{SSE}
Inspired by LD-based FSL methods \cite{DN4,ADM,DeepEMD}, we employ plenty of local descriptors (LDs) to measure the \emph{image-to-class} similarity. Considering spatial relationship of LDs, we believe that for two LDs that are spatially adjacent and therefore highly likely similar to each other, their similarity values to other LDs should be close and change in a smooth manner. However, the above methods did not effectively take this into account, and could generate noisy or non-smooth similarities with respect to the spatial dimension. 
Moreover, merely having a single similarity score as in the above methods cannot effectively serve the domain alignment required in our FS-UDA. Therefore, we propose a similarity encoding module to learn \emph{image-to-class} similarity patterns (SPs), by leveraging Gaussian filter and max-pooling to smooth and aggregate the similarities of LDs. The learning process of SPs is cast into three following parts: 

{\textbf{(1) Computing the similarity matrix of LDs (see step 1 in Figure  \ref{fig1}) as in DN4 \cite{DN4}.}} We firstly extract the LDs of a \emph{query image} $q$ and a \emph{support class} $c\ (c=1,..,N)$, and then construct their LD matrix $L_q^{que}\in\mathbb{R}^{HW \times C}$ and $L_c^{sup}\in \mathbb{R}^{KHW \times C}$, respectively, where each row represents a LD vector. $H,\ W$ and $C$ denote the height, the width and the number of channels of the feature map, respectively. Then, we compute their \emph{cosine} similarity matrix $M^c_q\in\mathbb{R}^{HW\times KHW}$ 
by $cos(L_q^{que},L_c^{sup})$. Finally, we only maintain the top-$k$ similarity values for each row in $M^c_q$ and set the remaining as zero.

\textbf{(2) Reshaping the similarity matrix $M^c_q$ into a 3D map (see step 2 in Figure \ref{fig1}).} To recover the spatial positions of \emph{query images} in the similarity matrix, we reshape the similarity matrix $M^c_q$ into a 3D similarity representation $M^{c'}_q\in\mathbb{R}^{H\times W\times KHW}$, by partitioning its first dimension $HW$ to the two dimensions $H$ and $W$.

\textbf{(3) Performing Gaussian filter and max-pooling to generate the SPs (see step 3 in Figure  \ref{fig1}).} According to our observation, several noises or non-smooth values exist on the first two-dimensional plane of $M^{c'}_q$. Considering that similarities of spatially-adjacent LDs to other LDs should be close, we conduct Gaussian filtering (with $3\times3\times 1$ size) from the first two dimensions of $M^{c'}_q$ to smooth their similarities. Then, max-pooling is employed to extract the discriminative similarity and finally generate the SP vectors $\{p^{c}_{q}\}$. Compared with a single similarity score used in \cite{DN4}, a SP contains the similarities of a \emph{query image} to all LDs of the \emph{support class}. Thus, the SP provides an informative and patterned measurement, which will be used for classification and domain alignment.

For classification, we calculate the \emph{image-to-class} similarity score for each \emph{support class} by summing all the elements of the SP vector, \emph{i.e.} $\mathbf{1}^{\top}\cdot p^{c}_{q}$, and then select the class label with the highest score as the predicted label of \emph{query image} $q$. The classification loss $\mathcal{L}_{cls}$ is calculated by cross-entropy loss, which can be written by:
\begin{equation}
\setlength\abovedisplayskip{0.5pt}
\setlength\belowdisplayskip{1pt}
    \mathcal{L}_{cls} = -\frac{1}{|Q_{\mathcal{S}}|}
    \sum_{q \in Q_{\mathcal{S}}} \log \frac{\exp(\mathbf{1}^{\top}\cdot p^{c^{*}}_{q})}{\sum_{c}\exp(\mathbf{1}^{\top}\cdot p^{c}_{q})},
\label{cls_loss}
\end{equation}
where $c^{*}$ is the actual label of the \emph{query image} $q$.

\subsection{Similarity Pattern Alignment}
To reduce the domain gap, we align the SPs of \emph{query images} from the source and target domains, because the SPs contain \emph{image-to-class} similarities that are crucial to final classification predictions. Considering that covariance matrix could exactly capture the underlying distribution information \cite{covnet,wass_covar,wass_covar2_began}, we align the covariance matrix of the SPs from both domains for domain adaptation.

For fine-grained alignment, we focus on the SP between a \emph{query image} $q$ and a specific \emph{support image} $i$ (namely \emph{image-to-image} SP, $p^i_q\in\mathbb{R}^{HW}$), instead of the \emph{image-to-class} one. In other words, the learned \emph{image-to-class} pattern $p^c_q\in\mathbb{R}^{KHW}$ can be decomposed into $K$ different \emph{image-to-image} patterns $p^i_q\in\mathbb{R}^{HW\times 1}, i=1,...,K$. 

Specifically, for a \emph{support image} $i$, we first collect its \emph{image-to-image} SPs from all the \emph{query images} in the source domain to form a set of SPs $\mathcal{E}^{i}_{\mathcal{S}}=\{p^{i}_{q}|q \in Q_{\mathcal{S}} \}$. Similarly, we also obtain the set of SPs in the target domain, \emph{i.e.} $\mathcal{E}^{i}_{\mathcal{T}}=\{p^{i}_{q}|q \in Q_{\mathcal{T}}\}$.
Then, the covariance matrix $\Sigma_{d}^{i}\in\mathbb{R}^{HW\times HW}$ for both domains ($d=\{\mathcal{S},\mathcal{T}\}$) can be calculated by
$\Sigma_{d}^{i}=\frac{1}{|\mathcal{E}^{i}_{d}|-1}\mathop{\sum}_{p_{q}^{i} \in \mathcal{E}^{i}_{d}} (p_{q}^{i} - \bar{p}^{i} )(p_{q}^{i} - \bar{p}^{i})^{\rm{T}}$, where $\bar{p}^{i}$ represents the averaged vector of SPs in the set $\mathcal{E}^{i}_{d}$. Finally, to align the two domains, we design a loss function of \emph{similarity pattern alignment (spa)} to minimize the difference between $\Sigma_{\mathcal{S}}^{i}$ and $\Sigma_{\mathcal{T}}^{i}$, which can be calculated as:
\begin{equation}
\setlength\abovedisplayskip{0.5pt}
\setlength\belowdisplayskip{1pt}
  \mathcal{L}_{spa}(\{\Sigma_{\mathcal{S}}^{i}\},\{\Sigma_{\mathcal{T}}^{i}\})=\frac{1}{NK} \sum _{i=1}^{NK}||\Sigma_{\mathcal{S}}^{i}-\Sigma_{\mathcal{T}}^{i}||_{F}^{2}, 
 \label{sea_loss}
\end{equation}
where $NK$ represents the number of images in \emph{N-way, K-shot} support set. By minimizing the loss $\mathcal{L}_{spa}$, the domain gap will be reduced. In some sense, this loss can also be viewed as a regularization of \emph{image-to-class} similarity measurement. This will be also validated in the experiments of few-shot learning.

\subsection{Local Descriptor Alignment}
\textbf{Domain Adversarial Module.} To align the domains on local features, we apply adversarial training on the LDs of \emph{query images} between $Q_{\mathcal{S}}$ and $Q_{\mathcal{T}}$, which are from the distributions $P_{\mathcal{S}}$ and $P_{\mathcal{T}}$, respectively. We design a domain discriminator $D$ to predict the domain label of each LD $l\in \mathbb{R}^{C}$. The adversarial training loss can be written by: 
\begin{equation}
\setlength\abovedisplayskip{1pt}
\setlength\belowdisplayskip{0.5pt}
    \min_F\max_D\mathcal{L}_{adv}=\mathop{\rm{E}}_{l\sim P_{\mathcal{S}}}[\log(1 -  D(l))] + \mathop{\rm{E}}_{l\sim P_{\mathcal{T}}}[\log D(l)]
\label{adv_loss}
\end{equation}
In each episode, we minimize the loss $\mathcal{L}_{adv}$ for the embedding network $F$ and maximize the loss for the discriminator $D$.

\textbf{Multi-scale Local Descriptor Matching.} In addition, to incorporate the discriminative information of LDs in the target domain for domain alignment, we develop a \emph{multi-scale local descriptor matching (msm)} module between domains. It is encouraged to push each LD $l$ in $Q_{\mathcal{T}}$ to its $k$ ($k>1$) nearest neighbors from multi-scale LDs in $X_{\mathcal{S}}$ in an unsupervised way. These multi-scale LDs are generated by average pooling on embedded feature maps with different scales ($5\times 5, 2\times 2, 1\times 1$), and \emph{cosine} similarity is utilized to find the nearest neighbors. This loss can be written as:
\begin{equation}
\setlength\abovedisplayskip{0.5pt}
\setlength\belowdisplayskip{1pt}
    \mathcal{L}_{msm} = -\frac{1}{|Q_{\mathcal{T}}|}
    \sum_{q \in Q_{\mathcal{T}}}\sum_{l \in q} \sum_{i=1}^{k}\log \frac{\exp(m_{l}^{i})}{\sum_{j=1}^{n}\exp(m_{l}^{j})},
\label{msm_loss}
\end{equation}
where $m_{l}^{j}\in [-1, 1]$ represents the \emph{cosine} similarity of the LD $l$ to its $j$-th neighbor in multi-scale LDs of $X_{\mathcal{S}}$. Also, for every \emph{query image}, we only measure the similarity of its $n$ neighbors, and the remaining neighbors are not considered. 

In sum, we combine the above losses to train the feature embedding network $F$ on a large amount of episodes. The overall objective function can be written as:
\begin{equation}
\setlength\abovedisplayskip{1.5pt}
\setlength\belowdisplayskip{1pt}
    \min_{F}\max_{D} [\mathcal{L}_{cls}+\lambda_{s}\mathcal{L}_{spa} +\lambda_{a}\mathcal{L}_{adv} + \lambda_{m}\mathcal{L}_{msm}],
\label{obj_func}
\end{equation}
where the parameters $\lambda_{s},\lambda_{a}$ and $\lambda_{m}$ are to balance the effects of different loss terms $\mathcal{L}_{spa},\mathcal{L}_{adv}$ and $\mathcal{L}_{msm}$, respectively. The whole process of the proposed IMSE is summarized in Algorithm \ref{alg:algorithm}.

\begin{algorithm}[h]
\caption{Proposed IMSE}\label{alg:algorithm}
\hspace{-0.6cm}\textbf{Training Input}:  An auxiliary dataset $D^{\text{aux}}$ including labeled\\
\hspace{-0.3cm}source and unlabeled target domain data, parameters $\lambda_s,\lambda_a,\lambda_m$\\
\hspace{-0.4cm}\textbf{Training Output}: Embedding network $F$ and discriminator $D$\\

\begin{algorithmic}[1]
\STATE Randomly initialize $F$ and $D$.
\WHILE{not converged}
\STATE Sample an episode $\{X_{\mathcal{S}}, \mathcal{Q}_{\mathcal{S}},\mathcal{Q}_{\mathcal{T}}\}$ from $D^{\text{aux}}$.\\
\STATE\textcolor{cyan}{\texttt{\# compute similarity patterns}}\\
\STATE Calculate similarity patterns $\{p_{q}^{c}|q \in \mathcal{Q}_{\mathcal{S}} \bigcup \mathcal{Q}_{\mathcal{T}},c=1,...,N\}$ using $F(X_{\mathcal{S}})$, $F(\mathcal{Q}_{\mathcal{S}})$ and $F(\mathcal{Q}_{\mathcal{T}})$.
\STATE\textcolor{cyan}{\texttt{\# compute classification loss and domain alignment losses}}\\
\STATE Compute classification loss $\mathcal{L}_{cls}$ for labeled query set $\mathcal{Q}_{\mathcal{S}}.$
\STATE Compute SP alignment loss $\mathcal{L}_{spa}$ via Eq. (\ref{sea_loss}).
\STATE Compute LD alignment losses $L_{adv},L_{msm}$ via Eqs. (\ref{adv_loss})-(\ref{msm_loss}).
\STATE\textcolor{cyan}{\texttt{\# update parameters}}\\
\STATE Optimize the $F$ and $D$ by minimizing the objective in Eq. (\ref{obj_func}). 
\ENDWHILE
\end{algorithmic}

\hspace{-0.5cm}\textbf{Testing Input}: New tasks and trained embedding network $F$\\
\textbf{Testing Output}: Predictions $\{\hat{y}_{q}\}$ in target domain of new tasks\\

\begin{algorithmic}[1]
\FOR{ $\{X_{\mathcal{S}},\mathcal{Q}_{\mathcal{T}} \}$ in new tasks} 
    \STATE \textcolor{cyan}{{\texttt{\#\ test on a new task}}}\\
    \STATE Compute similarity patterns $\{p_{q}^{c}|q \in \mathcal{Q}_{\mathcal{T}},c=1,...,N\}$ using $F(X_{\mathcal{S}})$ and $F(\mathcal{Q}_{\mathcal{T}})$.
    \STATE Predict the labels $\{\hat{y}_{q}\}$, $\hat{y}_{q}={\arg \max}_{c}\{\mathbf{1}^{\top}\cdot p^{c}_{q}|c=1,...,N\}$.
\ENDFOR
\end{algorithmic}
\end{algorithm}

\section{Experiments}
\textbf{\emph{DomainNet} dataset.} To demonstrate the efficacy of our method, we conduct extensive experiments on a multi-domain benchmark dataset \emph{DomainNet}. It was released in 2019 for the research of multi-source domain adaptation \cite{domainnet}. It contains 345 categories and six domains per category, \emph{i.e.} \emph{quickdraw, clipart, real, sketch, painting, infograph} domains. In our experiments, we first remove the data-insufficient domain \emph{infograph}, and select any two from the remaining five domains as the source and target domains. Thus, we have 20 combinations totally for evaluation. Then, we discard the 37 categories that contain less than 20 images for the five domains. Finally, we randomly split the images of the remaining 308 categories into the images of 217 categories, 43 categories and 48 categories for episodic training (forming the auxiliary dataset), model validation and testing new tasks, respectively. Note that in each split every category contains the images of the five domains.

\textbf{Network architecture.} We employ ResNet-12 as the backbone of the embedding network, which is widely used in few-shot learning \cite{meta-baseline}. As for similarity pattern encoder, we set a $3\times 3 \times1$ \emph{Gaussian kernel} with standard deviation $0.8$ and stride $1$. Then, a $2\times 2$ max-pooling with stride $2$ is conducted. As for adversarial training, we use three fully-connected layers to build a domain discriminator. 

\textbf{Model training.} Before episodic training, inspired by \cite{meta-baseline}, a pre-trained the embedding network is trained by source domain data in the auxiliary dataset, which is used for performance improvement. Specifically, we first train the ResNet-12 with a full-connected layer to classify the source domain images in the auxiliary dataset. After the pre-training, we remove the full-connected layer and retain the convolutional blocks of ResNet-12 as the initialized model of the episodic training. Afterwards, we perform episodic training on 10000 episodes. For each episode, we randomly sample a support set (including 5 categories, and 1 or 5 source domain samples per category) and a query set (including 15 source domain samples for each of the same 5 categories and 75 randomly sampled target domain samples). 
During episode training, the total loss in Eq. (\ref{obj_func}) is minimized to optimize the network parameters for each episode. Also, we employ Adam optimizer with an initial learning rate of $1e\text{-}4$, and reduce the learning rate by half every 1000 episodes. 

\textbf{Model validation and testing.} For model validation, we compare the performance of different model parameters on 3000 tasks, which is randomly sampled from the validate set containing 43 categories. Then, we select the model parameters with the best validation accuracy for testing. During the testing phase, we randomly construct 3000 new FS-UDA tasks from the testing set containing 48 categories. For each task, the support set includes 5 categories, and 1 or 5 source domain samples per category, and the query set only includes 75 target domain samples from the same 5 categories for evaluation. The averaged top-1 accuracy on the 3000 tasks is calculated as the evaluation criterion.

\begin{table*}
\renewcommand{\arraystretch}{1.1}
\setlength{\tabcolsep}{6pt}
\caption{\small{Comparison of our method with the UDA, FSL and their combination methods for 5-way 1-shot or 5-shot FS-UDA tasks. Each row represents the accuracy (\%) of a compared method adapting between two domains, where the \emph{skt, rel, qdr, pnt, and cli} denote the \emph{sketch, real, quickdraw, painting, and clipart} domains in the \emph{DomainNet} dataset, respectively. The best results are in bold.}}
\vspace{-0.4cm}
\resizebox{\linewidth}{!}{
\begin{tabular}{c|cccccccccc|c}
\hline\hline
\multicolumn{11}{c}{\textbf{5-way, 1-shot}} \\
\hline \multirow{2}{*}{\textbf{Methods}}
    &\textbf{skt$\leftrightarrow$rel} &\textbf{skt$\leftrightarrow$qdr} &\textbf{skt$\leftrightarrow$pnt} &\textbf{skt$\leftrightarrow$cli} &\textbf{rel$\leftrightarrow$qdr} &\textbf{rel$\leftrightarrow$pnt} &\textbf{rel$\leftrightarrow$cli} &\textbf{qdr$\leftrightarrow$pnt} &\textbf{qdr$\leftrightarrow$cli} &\textbf{pnt$\leftrightarrow$cli} & \textbf{avg}\\
     &$\rightarrow$ / $\leftarrow$&$\rightarrow$ / $\leftarrow$&$\rightarrow$ / $\leftarrow$&$\rightarrow$ / $\leftarrow$&$\rightarrow$ / $\leftarrow$&$\rightarrow$ / $\leftarrow$&$\rightarrow$ / $\leftarrow$&$\rightarrow$ / $\leftarrow$&$\rightarrow$ / $\leftarrow$&$\rightarrow$ / $\leftarrow$ & -- \\\hline
\textbf{MCD} \cite{MCD}
    &48.07/37.74 &38.90/34.51 &39.31/35.59 &51.43/38.98 &24.17/29.85 &43.36/47.32 &44.71/45.68 &26.14/25.02 &42.00/34.69 &39.49/37.28 &38.21\\
\textbf{ADDA} \cite{ADDA}
    &48.82/46.06 &38.42/40.43 &42.52/39.88 &50.67/47.16 &31.78/35.47 &43.93/45.51 &46.30/47.66 &26.57/27.46 &46.51/32.19 &39.76/41.24 &40.91\\
\textbf{DWT} \cite{DWT}
    &49.43/38.67 &40.94/38.00 &44.73/39.24 &52.02/50.69 &29.82/29.99 &45.81/50.10 &52.43/51.55 &24.33/25.90 &41.47/39.56 &42.55/40.52 &41.38\\
\hline
\textbf{ProtoNet} \cite{protonet}
    &50.48/43.15   &41.20/32.63 &46.33/39.69 &53.45/48.17 &32.48/25.06 &49.06/50.30   &49.98/51.95 &22.55/28.76   &36.93/40.98 &40.13/41.10 &41.21\\
\textbf{DN4} \cite{DN4}
     &52.42/47.29   &41.46/35.24   &46.64/46.55   &54.10/51.25 &33.41/27.48   &52.90/53.24   &53.84/52.84 &22.82/29.11   &36.88/43.61 &47.42/43.81 &43.61\\
\textbf{ADM} \cite{ADM}
      &49.36/42.27 &40.45/30.14 &42.62/36.93 &51.34/46.64 &32.77/24.30 &45.13/51.37 &46.8/50.15 &21.43/30.12 &35.64/43.33 &41.49/40.02 
        &40.11 \\
\textbf{FEAT} \cite{FEAT}
     &51.72/45.66   &40.29/35.45   &47.09/42.99   &53.69/50.59  &33.81/27.58   &52.74/53.82   &53.21/53.31 &23.10/29.39   &37.27/42.54 &44.15/44.49 &43.14\\
\textbf{DeepEMD} \cite{DeepEMD}
     &52.24/46.84   &42.12/34.77   &46.64/43.89   &55.10/49.56 &34.28/28.02   &52.73/53.26   &54.25/54.91 &22.86/28.79   &37.65/42.92 &44.11/44.38 &43.46\\
\hline
\textbf{ADDA+ProtoNet}
    &51.30/43.43 &41.79/35.40 &46.02/41.40 &52.68/48.91 &37.28/27.68 &50.04/49.68 &49.83/52.58 &23.72/32.03 &38.54/44.14 &41.06/41.59 &42.45 \\
\textbf{ADDA+DN4}
    &53.04/46.08 &42.64/36.46 &46.38/\textbf{47.08} &54.97/51.28 &34.80/29.84 &	\textbf{53.09}/54.05 &54.81/55.08 &23.67/31.62 &42.24/45.24 &46.25/44.40 &44.65\\
\textbf{ADDA+ADM}
      &51.87/45.08  &43.91/32.38  &47.48/43.37  &54.81/51.14  &35.86/28.15  &48.88/51.61  &49.95/54.29  &23.95/33.30  &43.59/48.21  &43.52/43.83 & 43.76\\
\textbf{ADDA+FEAT}
    &52.72/46.08 &47.00/36.94 &47.77/45.01 &56.77/52.10 &36.32/30.50 &49.14/52.36 &52.91/53.86 &24.76/35.38 &44.66/48.82 &45.03/45.92 &45.20\\
\textbf{ADDA+DeepEMD}
    &53.98/47.55 &44.64/36.19 &46.29/45.14 &55.93/50.45  &37.47/30.14 &52.21/53.32 &54.86/54.80  &23.46/32.89 &39.06/46.76  &45.39/44.65 &44.75\\
\hline
\textbf{IMSE (ours)}
     &\textbf{57.21}/\textbf{51.30} &\textbf{49.71}/\textbf{40.91} &\textbf{50.36}/46.35 &\textbf{59.44}/\textbf{54.06} &\textbf{44.43}/\textbf{36.55} &52.98/\textbf{55.06} &\textbf{57.09}/\textbf{57.98} &\textbf{30.73}/\textbf{38.70} &\textbf{48.94}/\textbf{51.47} &\textbf{47.42}/\textbf{46.52} &\textbf{48.86}\\
\hline\hline
\multicolumn{11}{c}{\textbf{5-way, 5-shot}} \\
\hline
\textbf{MCD} \cite{MCD}
    &66.42/47.73 &51.84/39.73 &54.63/47.75 &72.17/53.23 &28.02/33.98 &55.74/66.43 &56.80/63.07 &28.71/29.17 &50.46/45.02 &53.99/48.24 &49.65\\
\textbf{ADDA} \cite{ADDA}
    &66.46/56.66 &51.37/42.33 &56.61/53.95 &69.57/65.81 &35.94/36.87 &58.11/63.56 &59.16/65.77 &23.16/33.50 &41.94/43.40 &55.21/55.86 &51.76\\
\textbf{DWT} \cite{DWT}
    &67.75/54.85 &48.59/40.98 &55.40/50.64 &69.87/59.33 &36.19/36.45 &60.26/68.72 &62.92/67.28 &22.64/32.34 &47.88/50.47 &49.76/52.52 &51.74\\
\hline
\textbf{ProtoNet} \cite{protonet}
     &65.07/56.21  &52.65/39.75  &55.13/52.77   &65.43/62.62  
     &37.77/31.01   &61.73/66.85   &63.52/66.45  
     &20.74/30.55  &45.49/55.86 
     &53.60/52.92 &51.80\\
\textbf{DN4} \cite{DN4}
     &63.89/51.96   &48.23/38.68   &52.57/51.62   &62.88/58.33 &37.25/29.56   &58.03/64.72   &61.10/62.25 &23.86/33.03   &41.77/49.46 &50.63/48.56 &49.41\\
\textbf{ADM} \cite{ADM}
    &66.25/54.20 &53.15/35.69 &57.39/55.60 &71.73/63.42 &44.61/24.83 &59.48/69.17 &62.54/67.39 &21.13/38.83 &42.74/58.36 &56.34/52.83 &52.78\\
\textbf{FEAT} \cite{FEAT}
     &67.91/58.56 &52.27/40.97 &59.01/55.44 &69.37/65.95 &40.71/28.65 &63.85/71.25 &65.76/68.96 &23.73/34.02 &42.84/53.56 &57.95/54.84 &53.78\\
\textbf{DeepEMD} \cite{DeepEMD}
     &67.96/58.11 &53.34/39.70 &59.31/56.60 &70.56/64.60 &39.70/29.95 &62.99/70.93 &65.07/69.06 &23.86/34.34 &45.48/53.93 &57.60/55.61 &53.93\\
\hline

\textbf{ADDA+ProtoNet}
    &66.11/58.72 &52.92/43.60 &57.23/53.90 &68.44/61.84 &45.59/38.77 &60.94/69.47 &66.30/66.10 &25.45/41.30 &46.67/56.22 &58.20/52.65 &54.52\\
\textbf{ADDA+DN4}
    &63.40/52.40 &48.37/40.12 &53.51/49.69 &64.93/58.39 &36.92/31.03 &57.08/65.92 &60.74/63.13 &25.36/34.23 &48.52/51.19 &52.16/49.62 &50.33\\
\textbf{ADDA+ADM}
      &64.64/54.65  &52.56/33.42  &56.33/54.85  &70.70/63.57  &39.93/27.17  &58.63/68.70  &61.96/67.29  &21.91/39.12  &41.96/59.03  &55.57/53.39 & 52.27\\
\textbf{ADDA+FEAT}
    &67.80/56.71 &60.33/43.34 &57.32/58.08 &70.06/64.57 &44.13/35.62 &62.09/70.32 &57.46/67.77 &29.08/44.15 &49.62/63.38 &57.34/52.13 &55.56\\

\textbf{ADDA+DeepEMD}
    &68.52/59.28 &56.78/40.03 &58.18/57.86 &70.83/65.39 
    &42.63/32.18 &63.82/\textbf{71.54} &66.51/69.21
    &26.89/42.33 &47.00/57.92 
    &57.81/55.23 &55.49\\
\hline
\textbf{IMSE (ours)}
    &\textbf{70.46}/\textbf{61.09} &\textbf{61.57}/\textbf{46.86} &\textbf{62.30}/\textbf{59.15} &\textbf{76.13}/\textbf{67.27} &\textbf{53.07}/\textbf{40.17} &\textbf{64.41}/70.63 &\textbf{67.60}/\textbf{71.76} &\textbf{33.44}/\textbf{48.89} &\textbf{53.38}/\textbf{65.90} &\textbf{61.28}/\textbf{56.74} &\textbf{59.60}\\
\hline\hline
\end{tabular}
}
\label{comparison}
\vspace{-0.3cm}
\end{table*}

\begin{table*}[]
\centering
\caption{\small{Ablation study of Gaussian kernel and max-pooling used in our IMSE to generate SPs.}}\label{abl_encoder}
\vspace{-0.4cm}
\resizebox{1\linewidth}{!}{
\begin{tabular}{cc|cccccccccc|c}
\hline\hline
\multirow{2}{*}{\textbf{Filters}}&\multirow{2}{*}{\textbf{Poolings}}
    &skt$\leftrightarrow$rel &skt$\leftrightarrow$qdr &skt$\leftrightarrow$pnt &skt$\leftrightarrow$cli &rel$\leftrightarrow$qdr &rel$\leftrightarrow$pnt &rel$\leftrightarrow$cli &qdr$\leftrightarrow$pnt &qdr$\leftrightarrow$cli &pnt$\leftrightarrow$cli & avg \\
     & &$\rightarrow$ / $\leftarrow$&$\rightarrow$ / $\leftarrow$&$\rightarrow$ / $\leftarrow$&$\rightarrow$ / $\leftarrow$&$\rightarrow$ / $\leftarrow$&$\rightarrow$ / $\leftarrow$&$\rightarrow$ / $\leftarrow$&$\rightarrow$ / $\leftarrow$&$\rightarrow$ / $\leftarrow$&$\rightarrow$ / $\leftarrow$ & -- \\
\hline
\textbf{Gaussian}&\textbf{MaxPooling}&\textbf{57.21}/\textbf{51.30} &\textbf{49.71}/\textbf{40.91} &\textbf{50.36}/\textbf{46.35} &\textbf{59.44}/\textbf{54.06} &\textbf{44.43}/\textbf{36.55} &\textbf{52.98}/\textbf{55.06} &\textbf{57.09}/\textbf{57.98} &\textbf{30.73}/\textbf{38.70} &\textbf{48.94}/\textbf{51.47} &\textbf{47.42}/\textbf{46.52} &\textbf{48.86}\\
\textbf{Gaussian}&\textbf{DownSampling}&50.01/42.72 &34.11/30.62 &37.71/33.47 &51.91/43.78 &36.22/26.30 &41.18/45.33 &49.41/48.43 &23.61/31.30 &36.06/43.63 &35.40/33.73 &38.74\\
\textbf{Average}&\textbf{MaxPooling}&40.81/40.23 &37.54/28.81 &35.51/32.93 &40.10/37.16 &34.79/24.62 &37.73/40.35 &41.32/40.91 &21.48/30.32 &34.74/44.50 &33.16/31.97 &35.44\\
\textbf{None}&\textbf{MaxPooling}&34.48/30.24 &26.32/24.48 &27.62/30.39 &32.87/28.86 &28.76/22.73 &31.82/34.54 &31.39/31.45 &21.20/25.57 &24.82/29.72 &29.22/26.95 &28.67\\
\hline\hline
\end{tabular}
}
\vspace{-0.4cm}
\end{table*}

\subsection{Comparison Experiments for FS-UDA}
We conduct extensive experiments on \emph{DomainNet} to compare our method with related FSL, UDA, and their combination methods. 

\textbf{UDA methods.} We choose MCD \cite{MCD}, ADDA \cite{ADDA} and DWT \cite{DWT} to compare with our method. Specifically, 1) we first train the UDA models by using all training categories in the auxiliary set; 2) Secondly, we replace the last classification layer of the models with a new \emph{5-way} classifier for various tasks, as in \emph{baseline++} \cite{jiabin2}; 3) Thirdly, for testing on each task, we freeze the trained backbone and only fine-tune the classifier for a novel task with its support set, by using an SGD optimizer (learning rate $0.01$, momentum $0.9$ and weight decay $1e\text{-}4$) with batch size $4$ for $300$ iterations. Finally, we test the classification performance on its query set.

\textbf{FSL methods.} Since our method is related to metric-based FSL methods, we choose five representative metric-based methods, \emph{i.e.} ProtoNet \cite{protonet}, DN4 \cite{DN4}, ADM \cite{ADM}, FEAT \cite{FEAT}, and DeepEMD \cite{DeepEMD} for comparisons. Especially, DN4, ADM and DeepEMD also utilize local descriptors for classification. For fair comparison, we also pre-train the embedding network and perform episodic training on auxiliary dataset \cite{meta-baseline}. Note that, FSL methods don't handle the domain shift between the support and query sets, and are also not capable of leveraging the target domain data in the auxiliary dataset because of no labels in the target domain. Thus, we further combine the FSL and UDA methods for domain adaptation and classification.

\textbf{FSL methods combined with ADDA \cite{ADDA}.} We combine the five above FSL methods with ADDA \cite{ADDA}, which are abbreviated as ADDA+ProtoNet, ADDA+DN4, ADDA+ADM, ADDA+FEAT, and ADDA+DeepEMD, respectively. Specifically, we add the feature-level adversarial loss in ADDA \cite{ADDA} for domain alignment in every episode.
The used discriminator is the same with our method. For fair comparison, these combination methods also pretrain the embedding network before episodic training.

\textbf{Comparison Analysis.} Table \ref{comparison} shows the results of all the compared methods for 20 cross-domain combinations, which records the averaged classification accuracy of target domain samples over 3000 \emph{5-way 1-shot/5-shot} UDA tasks. As reported in Table \ref{comparison}, our IMSE achieves the best performance for the most combinations and their average. Specifically, the UDA and FSL baselines in the first two parts of Table \ref{comparison} perform much worse than our method. This is because the UDA models did not consider task-level generalization, and the FSL models are lack of handling domain shift. In the third part of Table \ref{comparison}, the combination methods we build perform domain adversarial training over every episode. They are thus generally better than the above two parts, but still inferior to our IMSE. This is because the combination methods with ADDA \cite{ADDA} only perform the domain alignment based on feature maps, not considering the alignment of both local descriptors (more fine-grained than feature maps) and similarity patterns (related to classification predictions) used in our IMSE. This shows the efficacy of our IMSE for FS-UDA.

On the other hand, we can see that the 20 cross-domain combinations have considerably different performances. This is because several domains (\emph{e.g. quickdraw}) are significantly different from other domains, while some other domains (\emph{e.g. real, painting}) are with the similar styles and features. Thus, the performance becomes relatively low when large domain gap is presented, and the situation will become better when the domain gap reduces. For example, from \emph{quickdraw} to \emph{painting/real}, most methods perform worse due to larger domain gap, but our IMSE and ADDA \cite{ADDA} perform much better than the others, showing the positive effect of their domain alignment. From \emph{painting} to \emph{real}
and \emph{real} to \emph{painting} which have smaller domain gap, the FSL, combinational methods and our IMSE perform better than the UDA methods, showing the positive effect of task generalization. In general, this reflects the advantages of our IMSE to deal with domain shift and task generalization in FS-UDA, especially for large domain gap.

\begin{figure}
\begin{center}
    \includegraphics[width=0.96\linewidth]{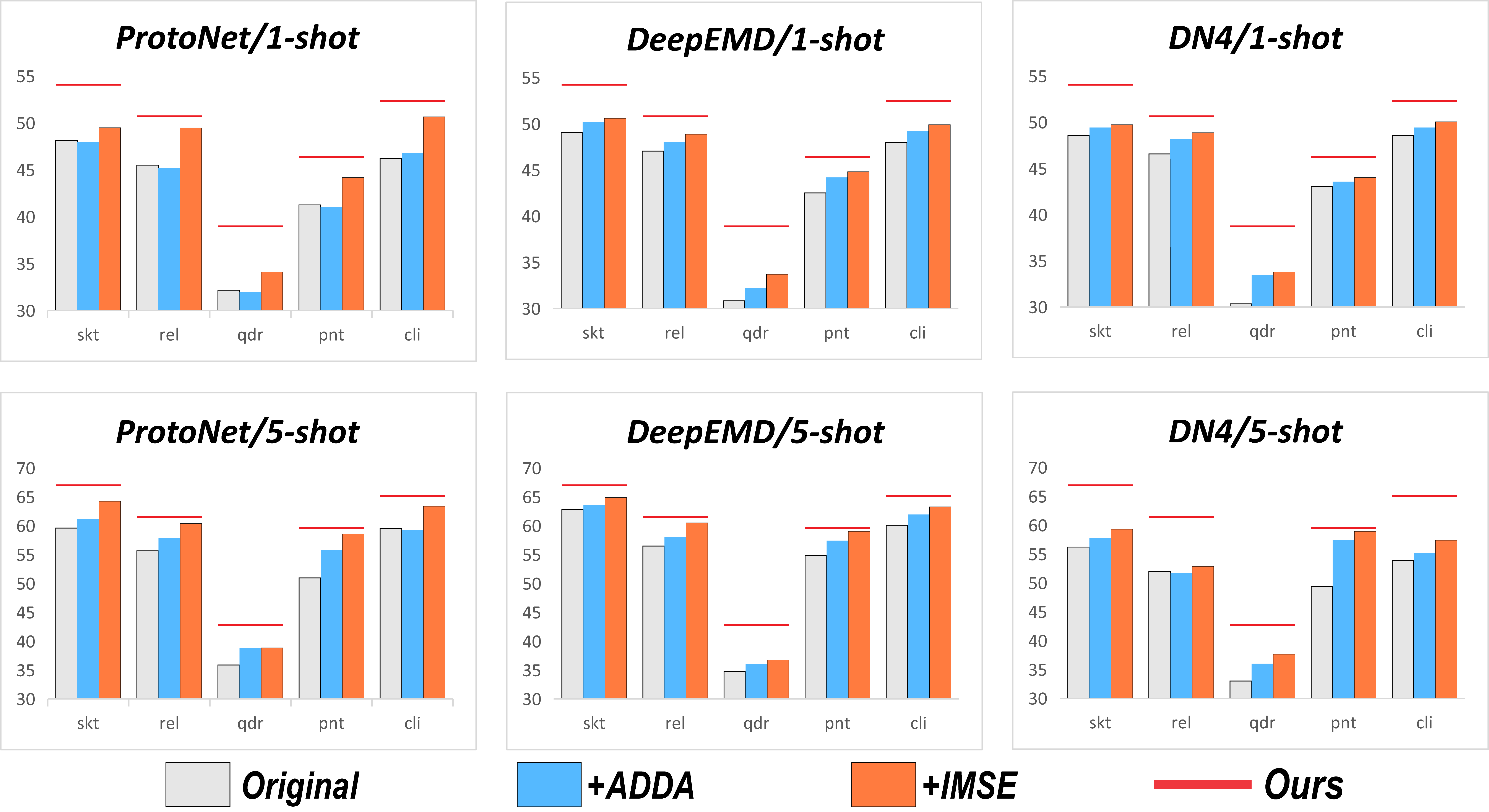}
\end{center}
\vspace{-0.4cm}
\caption{\small{Comparison of applying different domain alignment modules to three FSL methods ProtoNet, DeepEMD and DN4. The red lines denote the accuracy of our method. }}
\label{imse_improve}
\vspace{-0.4cm}
\end{figure}

\textbf{Evaluation on FSL methods with our domain alignment modules.} To further validate the efficacy of the two alignment modules designed in our IMSE, \emph{i.e.} similarity pattern alignment and local descriptor alignment, we incorporate their corresponding loss terms $\mathcal{L}_{spa}$, $\mathcal{L}_{adv}$ and $\mathcal{L}_{msm}$ into the loss functions of three FSL models \emph{ProtoNet}, \emph{DeepEMD} and \emph{DN4}, namely \emph{IMSE+ProtoNet}, \emph{IMSE+DeepEMD} and \emph{IMSE+DN4}, respectively. We test them on 3000 new FS-UDA tasks, compared with the FSL methods (\emph{ProtoNet}, \emph{DeepEMD} and \emph{DN4}) and the combination methods with ADDA \cite{ADDA} (\emph{ADDA+ProtoNet}, \emph{ADDA+DeepEMD} and \emph{ADDA+DN4}). For simplification and clarification, we average the accuracy from every domain to the other four domains, which are shown in Figure \ref{imse_improve}. Obviously, these methods \emph{IMSE+ProtoNet}, \emph{IMSE+DeepEMD} and \emph{IMSE+DN4} are more significantly improved by using our alignment modules in IMSE, which indicates the effectiveness of aligning the SPs and LDs again. Moreover, these methods are still outperformed by our IMSE, because we have a complete and self-contained FS-UDA framework, and the designed similarity patterns are effective for classification. This shows the efficacy of our IMSE again.

\textbf{Different losses.} We conduct various experiments on \emph{DomainNet} to further evaluate the effect of our different losses in Eq.(\ref{obj_func}). Besides the classification loss ($\mathcal{L}_{cls}$), we combine the remaining three loss terms: 1) LD adversarial training loss ($\mathcal{L}_{adv}$), 2) multi-scale LD matching loss ($\mathcal{L}_{msm}$), and 3) similarity pattern alignment loss ($\mathcal{L}_{spa}$). We evaluate both 1-shot and 5-shot FS-UDA tasks by taking \emph{sketch} as the source domain, and the other four domains as the target domains, respectively. The accuracies on the four target domains are reported in Table \ref{ablation}. As observed, the more the number of loss terms involved, the higher the accuracy. The combination of all the three losses is the best. For the single loss, $\mathcal{L}_{spa}$ performs better than $\mathcal{L}_{adv}$ and $\mathcal{L}_{msm}$, showing the efficacy of SP alignment. The combination of $\mathcal{L}_{adv}$ and $\mathcal{L}_{msm}$ is considerably better than either one of them, showing the positive effect of combining multi-scale LD matching and LD adversarial training. Based on the above, the addition of $\mathcal{L}_{spa}$ further improves the performance, indicating the necessity of jointly aligning SPs and LDs.

\textbf{Gaussian kernel and maxpooling for SPs.} We further investigate the process of generating our SPs by applying Gaussian kernel and maxpooling (see Figure \ref{fig1}). We try to change the filters and the poolings, and compare the variants in Table \ref{abl_encoder}. As seen, our method in the first row performs best. The performance decreases sharply when Gaussian filter is replaced or removed (see rows 3-4 in Table \ref{abl_encoder}). This is because Gaussian kernel efficiently makes the similarity of adjacent points smooth. For downsampling in the second row, we remove the maxpooling and perform the downsampling by changing the filter stride to 2, but it performs worse than our IMSE using maxpooling in the first row. This is because maxpooling extracts local maximum responses of similarity maps that could reflect the prominent and discriminative characteristics for classification.

\begin{table}
\small
\begin{center}
\caption{\small{Ablation study of the three losses designed in our IMSE, where the FS-UDA tasks are evaluated from a source domain (\emph{skt}) to four different target domains (\emph{rel, qdr, pnt,} and \emph{cli}) in \emph{DomainNet}.}}\label{ablation}
\vspace{-0.45cm}
\renewcommand{\arraystretch}{0.88}
\setlength{\tabcolsep}{8pt}
\begin{tabular}{ccc|cccc}
\hline\hline
\multicolumn{3}{c|}{\textbf{Components}}         &\multicolumn{4}{c}{\textbf{Target Domains}}  \\
\hline
$\mathcal{L}_{adv}$ &$\mathcal{L}_{msm}$ &$\mathcal{L}_{spa}$ &\emph{rel} &\emph{qdr} &\emph{pnt} &\emph{cli} \\
\hline
\          &\          &\           &54.42 &42.28 &46.80 &56.37 \\
\checkmark &\          &\           &52.11 &44.51 &48.76 &56.99 \\
\          &\checkmark &\           &54.53 &42.10 &47.62 &56.58 \\
\          &\          &\checkmark  &56.12 &48.37 &49.44 &58.78 \\
\checkmark &\checkmark &\           &55.30 &45.52 &49.53 &58.34 \\
\checkmark &\          &\checkmark  &56.57 &49.03 &50.17 &59.12 \\
\          &\checkmark &\checkmark  &56.00 &47.93 &50.13 &58.33 \\
\checkmark &\checkmark &\checkmark  &\textbf{57.21}   &\textbf{49.71}     &\textbf{50.36}   &\textbf{59.30}\\
\hline\hline
\end{tabular}
\end{center}
\vspace{-0.4cm}
\end{table}

\begin{figure}[t]
\vspace{-0.3cm}
\begin{minipage}[b]{0.39\columnwidth}
\centering
\includegraphics[width=0.94\columnwidth]{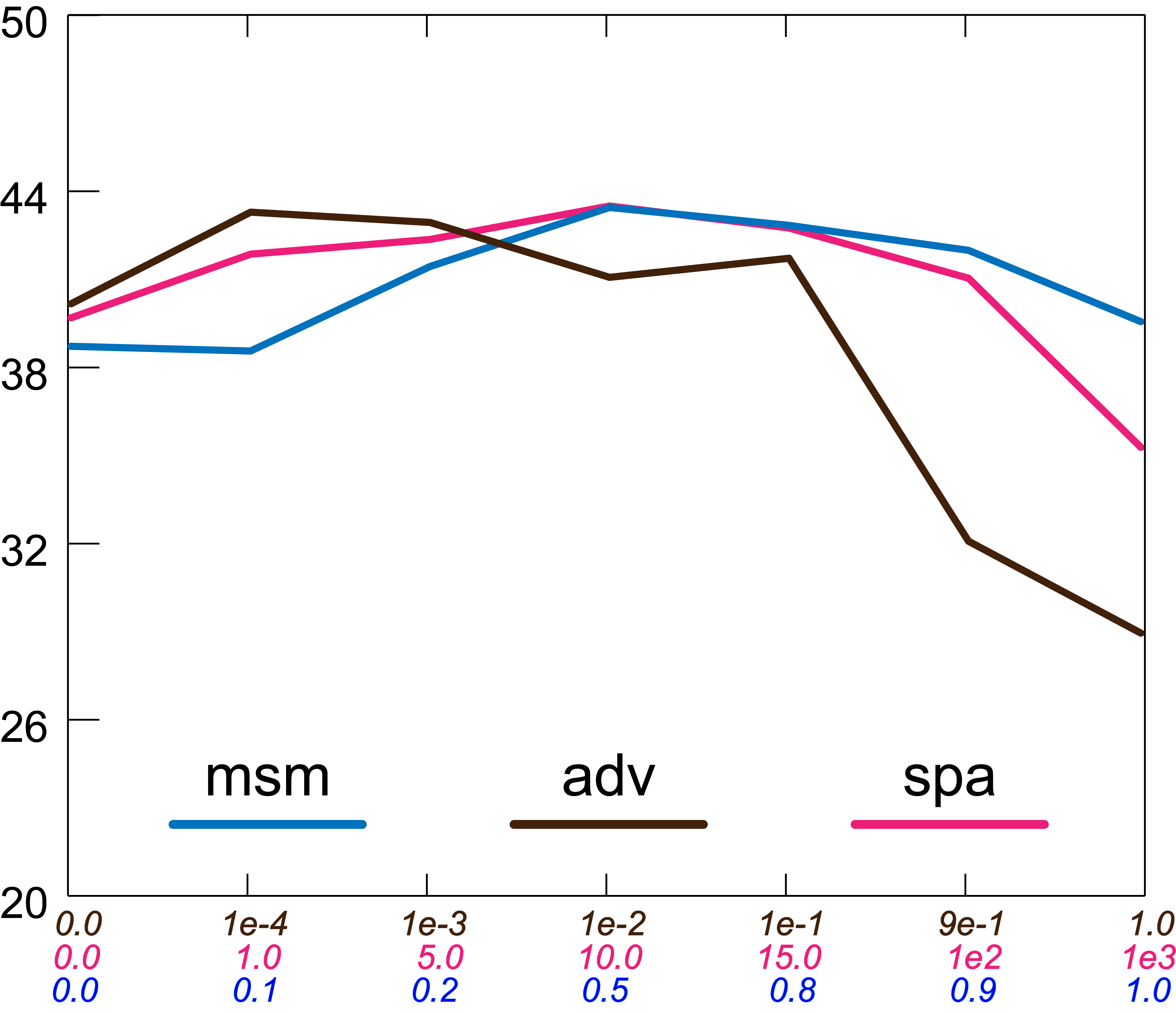}
\vspace{-0.4cm}
\caption{\small{Sensitivity of $\lambda_a$, $\lambda_s$ and $\lambda_m$ for \emph{rel}$\rightarrow$\emph{qdr}.}}\label{sensitive}
\vspace{-0.8cm}
\end{minipage}
\begin{minipage}[b]{0.6\columnwidth}
\small
\centering
\makeatletter\def\@captype{table}\makeatother
\caption{\small{Cross-dataset evaluation for \emph{5-way 1-shot} UDA tasks between two domains \emph{rel} and \emph{skl}: performing episodic training on \emph{DomainNet}, and testing on expanded dataset \emph{miniImageNet}*.}}\label{crossdataset}
\setlength{\tabcolsep}{3.3pt}
\renewcommand{\arraystretch}{0.96}
\begin{tabular}{c|c|c}
\hline\hline
\textbf{Methods}&\emph{skl}$\rightarrow$\emph{rel} & \emph{rel}$\rightarrow$\emph{skl}\\\hline
\textbf{DN4}    &44.01$\pm$0.87 & 40.61$\pm$0.90\\
\textbf{DeepEMD}&46.14$\pm$0.82 & 45.91$\pm$0.77\\
\textbf{IMSE (ours)}&\textbf{48.78}$\pm$0.78&\textbf{48.52}$\pm$0.81\\\hline\hline
\end{tabular}
\end{minipage}
\vspace{-0.3cm}
\end{figure}

\textbf{Cross-dataset evaluation.} We investigate the performance of our model on a substantially different dataset \emph{miniImageNet}, after episodic training on \emph{DomainNet}. To produce two domains for FS-UDA, we modify \emph{miniImageNet} by transferring a half of real images (\emph{rel}) into sketch images (\emph{skt}) by MUNIT \cite{MUNIT2018}. We conduct experiments between \emph{rel} and \emph{skt}, and compare our IMSE with DN4 and DeepEMD for \emph{5-way 1-shot} UDA tasks. Their accuracies are shown in Table \ref{crossdataset}. As seen, our method still performs better than DN4 and DeepEMD, showing the efficacy of our IMSE again. Compared with \emph{DomainNet}, the results in \emph{miniImageNet} have some performance degeneration, since their sketch images are more simplified than that in \emph{DomainNet}, and thus have larger difference from real images.

\textbf{Hyper-parameters sensitive analysis.} To investigate the effect of three domain alignment losses, we conduct the sensitive experiments of three hyper-parameters in Eq. (\ref{obj_func}), \emph{i.e.} $\lambda_a$, $\lambda_s$ and $\lambda_m$, respectively. We fix two parameters as the optimal values, and meanwhile change the other one for performance evaluation. The changing accuracy curves of the three parameters are shown in Figure \ref{sensitive}. Obviously, the curves of the parameters $\lambda_s$ and $\lambda_m$ are stably changing. When $\lambda_a$ is less than $0.1$, it also changes slowly. This shows the low sensitivity of the three parameters.

\begin{table}
\small
\centering
\caption{\small{Comparison with the FSL baselines using the backbone ResNet-12 on \emph{miniImageNet}. The top two  results are in bold.}}\label{fsl_cmp}
\vspace{-0.4cm}
\renewcommand{\arraystretch}{0.98}
\setlength{\tabcolsep}{10pt}
\begin{tabular}{c|c c}
\hline\hline
\textbf{Method} & \textbf{1-shot} & \textbf{5-shot} \\
\hline
\textbf{ProtoNet \cite{protonet}}  & 60.37 $\pm$ 0.83   &78.02 $\pm$ 0.57 \\
\textbf{MTL \cite{mtl}}  &61.20 $\pm$ 1.80   &75.50 $\pm$ 0.80 \\
\textbf{MatchNet \cite{matchingnet}}  & 63.08 $\pm$ 0.80   &75.99 $\pm$ 0.60 \\
\textbf{TADAM \cite{tadam}}  & 58.50 $\pm$ 0.30   &76.70 $\pm$ 0.30 \\

\textbf{DN4 \cite{DN4}}  & 61.57 $\pm$ 0.72   &71.40 $\pm$ 0.61 \\
\textbf{CAN \cite{CAN}}  & 63.85 $\pm$ 0.48   &79.44 $\pm$ 0.34 \\
\textbf{DeepEMD \cite{DeepEMD}}  & \textbf{65.91} $\pm$ 0.82   &\textbf{82.41} $\pm$ 0.56 \\
\hline
\textbf{IMSE w/o $\mathcal{L}_{rspa}$}  &62.46 $\pm$ 0.87   &78.44 $\pm$ 0.47 \\
\textbf{IMSE (ours)}  &\textbf{65.35} $\pm$ 0.42   &\textbf{80.02} $\pm$ 0.34 \\
\hline\hline
\end{tabular}
\vspace{-0.5cm}
\end{table}

\subsection{Few-shot learning on \emph{miniImageNet}}
Additionally, we investigate the performance of our IMSE for few-shot learning with \emph{1-shot} and \emph{5-shot setting} on a benchmark dataset \emph{miniImageNet}. We adopt the same data partitioning, training and testing setting as in \cite{DN4}. Since in few-shot learning the support and query sets are from the same domain, only one domain exists and no domain alignment is required in this experiment. Thus, we discard the LD-based domain alignment losses $\mathcal{L}_{adv}$ and $\mathcal{L}_{msm}$ used in our method, and retain the classification loss $\mathcal{L}_{cls}$. Meanwhile, we modify the similarity pattern (SP) alignment loss $\mathcal{L}_{spa}$ to constrain the covariance matrix of SPs to be an identity matrix:
$\mathcal{L}_{rspa}(\{\Sigma^{i}\})=\frac{1}{NK} \sum_{i=1}^{NK}||\Sigma^{i}-\lambda \cdot I_{HM}||_{F}^{2}$,
where $\lambda = \frac{1}{HW}tr(\Sigma^{i})$. It can be viewed as a regularizer to learn less biased SPs for classification.

We compare our method with seven representative FSL methods in Table \ref{fsl_cmp}. The results of almost all the compared methods are quoted from \cite{DeepEMD}, except CAN and DN4. The results of CAN are quoted from its original work \cite{CAN}. Since DN4 in \cite{DN4} did not use backbone ResNet-12 (our setting), we use its released code, the backbone ResNet-12 and pretraining to do experiments for a fair comparison. Note that, we compare the FC version of DeepEMD, because it has no data augmentation, neither ours. Also, we evaluate IMSE by using the loss $\mathcal{L}_{rspa}$ or not. As observed in Table \ref{fsl_cmp}, IMSE performs better than most of recent FSL methods, showing excellent performance of IMSE for FSL. Meanwhile, our method is slightly lower than DeepEMD, especially for \emph{5-shot}. This may be because DeepEMD needs to learn new prototypes for each testing \emph{5-shot} task, which could improve the performance but is more time-consuming than ours. In addition, in our IMSE, it is much better to use the $\mathcal{L}_{rspa}$ than to remove it, which indicates the positive effect of using covariance matrix of SPs as regularization.

\section{Conclusion}
In this paper, we propose a novel method IMSE for the new setting of FS-UDA. Our IMSE leverages local descriptors to design the similarity patterns to measure the similarity of \emph{query images} and \emph{support classes} from different domains. To learn a domain-shared embedding model for FS-UDA, we align the local descriptors by adversarial training and multi-scale matching, and align the covariance matrices of the similarity patterns. Extension experiments over \emph{DomainNet} show the efficacy of our IMSE for FS-UDA, and our method also has good performance for FSL on \emph{miniImageNet}.

\begin{acks}
Wanqi Yang and Ming Yang are supported by National Natural Science Foundation of China (Grant Nos. 62076135, 61876087). Lei Wang is supported by an Australian Research Council Discovery Project (No. DP200101289) funded by the Australian Government.
\end{acks}

\bibliographystyle{ACM-Reference-Format}
\balance
\bibliography{sample-base}

\end{document}